\documentclass[conference]{IEEEtran}
\IEEEoverridecommandlockouts

\usepackage{cite}
\usepackage{amsmath,amssymb,amsfonts}
\usepackage{algorithmic}
\usepackage{graphicx}
\usepackage{textcomp}
\usepackage{xcolor}
\usepackage{csquotes}
\usepackage[colorlinks=true,urlcolor=teal,
            linkcolor=teal,citecolor=teal]{hyperref}
\usepackage{orcidlink}
\usepackage{booktabs}
\usepackage{multirow}
\usepackage{siunitx}

\newcommand{\genemail}[2]{\href{#1}{#2}}
\newcommand{\grid}[2]{#1\(\times\)\!#2}

\begin{document}

\title{Quality Inspection of Printed Circuit Board Pin Insertion via Semantic Segmentation and Board-Level Feature Extraction\\
\thanks{This work was partially supported by the European Regional Development Fund (ERDF) and by the Free State of Bavaria as part of the project AIM-SMEs (Grant No. 2506-014-3.2), co-funded by the European Union; also by the High-Tech Agenda of the Free State of Bavaria.}
}

\newcommand{\othr}{\textit{Technical University of}\\\textit{Applied Sciences Regensburg}}

\author{\IEEEauthorblockN{Nils Rabeneck\orcidlink{0009-0005-4250-5490}}
\IEEEauthorblockA{\othr\\
Regensburg, Germany} 
\genemail{nils.rabeneck@othr.de}{nils.rabeneck@othr.de}
\and
\IEEEauthorblockN{André Kiunke}
\IEEEauthorblockA{\othr\\
Regensburg, Germany}
\genemail{andre.kiunke@st.othr.de}{andre.kiunke@st.othr.de}
\and
\IEEEauthorblockN{Nicole Hoess\orcidlink{0009-0002-6194-5887}}
\IEEEauthorblockA{\othr\\
Regensburg, Germany}
\genemail{nicole.hoess@othr.de}{nicole.hoess@othr.de}
\and
    \IEEEauthorblockN{Wolfgang Mauerer\orcidlink{0000-0002-9765-8313}}
    \IEEEauthorblockA{\othr\\
        \textit{Siemens AG, Technology} \\
        Regensburg/Munich, Germany \\
        \genemail{mailto:wolfgang.mauerer@othr.de}{wolfgang.mauerer@othr.de}
    }
}

\maketitle

\begin{abstract}
Quality control during printed circuit board (PCB) assembly is a critical step in ensuring reliable electronic products. Detecting misaligned pins during or after pin insertion remains a particularly challenging inspection task. This paper presents an automated defect detection method for identifying incorrectly inserted pins on PCBs. The proposed pipeline combines semantic segmentation using a U-Net architecture with contour-based feature extraction and logistic regression for board-level pass/fail classification. Segmentation masks are used to derive contour representations of individual pins, from which board-level features~—-~such as average contour size~—-~are extracted and used to train a logistic regression classifier.
We evaluate the method on two datasets: an industrial collection of real-world PCB images, and a publicly available PCB pin-inspection dataset with substantially different visual characteristics. To assess the effectiveness of the proposed approach, a comparison against PatchCore, an anomaly detection technique new to be applied to pin inspection, as well as instance segmentation-based pin detection is made. The developed method achieved Area Under the Receiver Operating Characteristic Curve (ROC-AUC) values of 0.990 on a random test set split from the industrial data and 1.000 on the public dataset indicating strong separation between pass and fail boards. The results indicate that the proposed approach is a promising candidate for automated pin inspection in industrial environments and achieves strong performance on datasets with substantially different visual characteristics after dataset-specific training.
\end{abstract}


\section{Introduction}

Image processing plays an important role in quality assessment for printed circuit boards (PCBs). It aims at removing defective PCBs identified from production at an early stage~\cite{anitha}. Advances in artificial intelligence (AI) methods for machine vision open up a wide range of possibilities for defect detection alongside established rule-based methods~\cite{chen}. Deep learning methods are regarded as established standard in PCB quality inspection for various defects because of their high accuracy and speed. However, they also face limitations, such as the range of defects, including open circuits, short circuits, under-etching, and spurious copper defects~\cite{ling}. While existing work focuses primarily on board-level defects and device defects~\cite{ural}, comparatively little work has addressed the detection and evaluation of PCB pins in industrial environments. However, pin defects can lead to quality issues and increased rejection rates during PCB manufacturing.

In this work, we consider pin insertion: the automated placement of pins on a PCB. Owing to the large number of insertion operations and tight manufacturing tolerances, defects such as misaligned pins can occur and may lead to quality issues, increased rejection rates, and costly rework~\cite{sixsigma}. Such defects occur for both, individual components and, more frequently, as systematic deviations of a desired process.
Our aim is to detect tilted pins directly after or even during pin insertion. Methods and datasets designed for component detection do not address the problem. Nevertheless, the successful implementation of AI-based methods for PCB inspection encourages the use of deep learning segmentation approaches for pin tilt detection. The primary goal is the development and evaluation of a decision function at the board-level, classifying PCB images as \emph{pass} or \emph{fail} based on detected pin misalignments. Since tilted pins are characterised by a larger surface area, semantic segmentation enables the extraction of contours and contour-based features at the board level.
Our main contributions are as follows:
\begin{enumerate}
    \item We investigate semantic segmentation for pin localisation in industrial PCB images.
    \item We propose a board-level defect classification pipeline based on contour features extracted from segmentation masks.
    \item We compare the proposed approach with alternative methods, including anomaly detection and object detection approaches.
    \item We evaluate the investigated methods on both an industrial dataset and a publicly available dataset and discuss their strengths and limitations for automated pin inspection.
    \item We provide a self-contained \href{https://github.com/lfd/paine2026}{reproduction package}~\cite{mauerer} that contains an environment allowing others to build directly upon our approach.
\end{enumerate}


\section{Related Work}
\label{sec:related-work}

A query in \href{https://www.webofscience.com/}{web of science} (July 18, 2026) for the terms \enquote{printed circuit board} AND (\enquote{artificial intelligence} OR \enquote{machine learning} OR \enquote{deep learning}) resulted in 362 publication results, with more than half  from 2024 or later. The majority of these references focuses on quality assurance using deep learning methods for the detection of mounted components and textual markings~\cite{ural,pineda}, or conductive structures on non-assembled PCBs~\cite{lu, deshmukh}. Some instance segmentation approaches for PCB-mounted components include pins as one of the target classes~\cite{jessurun, craig}. For component segmentation, YOLO~\cite{yolo} or ResNet~\cite{resnet} are frequently used convolutional neutral network (CNN) models~\cite{craig, ural}, while semantic segmentation models are also reported~\cite{pasunuri}. YOLO models were specifically studied for defect detection~\cite{aleman, lin}. In addition to supervised deep learning models, unsupervised methods have been considered suitable for PCB defect detection by comparing inspected boards to defect-free reference samples (\enquote{golden parts})~\cite{fridman}. There are several available annotated datasets regarding surface-mount device (SMD) detection, such as the FICS-PCB dataset~\cite{lu} and the FICS PCB Image Collection (FPIC) dataset~\cite{jessurun} with the aim of further development in the field of deep learning-based SMD localisation and defect recognition and are therefore not suitable with respect to pin detection on its own.  
For SMDs, the pin localisation method PinPoint~\cite{pinpoint} is known for quality assurance and assembly check. The algorithm processes contour data, but pin localisation is realised by a single point marking.

Regarding pin insertion and pin defect detection, traditional image processing methods such as threshold-based segmentation are still commonly applied, while many published approaches rely on specialised camera systems and fixed inspection conditions~\cite{tian}. In contrast, we develop an AI-based approach without focusing on a specific hardware setup. 
Pin inspection for insertion of through-hole components has been reported using instance segmentation of pins derived from 3D models or point clouds in a pre-insertion process~\cite{hagelskjaer}. Our method focuses on board-level defect detection after pin insertion and combines semantic segmentation with geometric feature extraction and statistical classification.
In addition to existing academic approaches, commercial machine-vision platforms are widely deployed in industrial PCB assembly. Systems such as those offered by \href{https://www.cognex.com/en/applications/assembly-inspection-and-verification/pcb-assembly-inspection-and-component-verification}{Cognex} support automated inspection tasks including component verification, placement inspection, and defect detection. However, reliable inspection of inserted pins remains challenging because of tight tolerances, component variability, and the occurrence of subtle alignment defects. Consequently, there is still a need for robust automated methods in pin-quality assessment.
CNN-based methods for pin insertion focus on the insertion process itself and a detection of tilted pin or pin defects is at most possible indirectly via sensory output~\cite{caporali} or implicitly by means of failed insertion~\cite{mou}. There is a publicly available \href{https://universe.roboflow.com/pcbpindetection/pin_pcb_detection}{data set} for PCB pin detection hosted by the Roboflow data platform with annotated data for YOLO instance segmentation~\cite{roboflow}. However, to the best of our knowledge, it is neither accompanied by a peer-reviewed publication nor has it been used in previously published academic work.

While the reviewed literature includes both classical machine-vision approaches and commercial inspection systems, many solutions rely on dedicated imaging setups or handcrafted inspection procedures. At the same time, CNN-based approaches predominantly focus on supporting the pin insertion process itself rather than on assessing the quality of the inserted pins. Consequently, the robust and automated detection of pin alignment defects remains insufficiently explored despite its relevance for industrial manufacturing practice. This motivates us to investigate and characterise semantic segmentation methods for pin-quality assessment in industrial PCB manufacturing environments.


\section{Method}
Pin alignment defects primarily are reflected by changes in the geometry and visible area of individual pins. Therefore, we perform semantic segmentation to obtain pixel-accurate representations of the pins. The resulting segmentation masks allow contour extraction and the computation of contour-based features, which form the basis for training a board-level decision function by means of logistic regression~\cite{bishop}.

\subsection{Datasets}

The first dataset was provided by an industrial partner and consists of 827 grey scale images of PCBs, each with a resolution \grid{4096}{3000} pixels. 759 of the images are without tilted pins, while all others show defective boards. Each image contains approximately 350 pins. Three pin geometries are present in the dataset. The majority are \enquote{standard pins}, while two additional types have increased pin width. Since the objective of the segmentation task is pin localisation rather than pin-type classification, all pin types were annotated as a single pin class.
Pins are arranged in grouped rows referred to as \emph{pin nests}, which are separated from neighbouring pin groups by a visible spacing. The images were taken in a laboratory setting reflecting an industrial environment. \SI{35}{\milli\metre} and \SI{50}{\milli\metre} camera lenses without perspective correction were used, and lightning conditions were varied to generate underexposed images and occasional strong reflections to resemble the demanding and changing conditions encountered in real-world production environments. Boards labelled as fail were manually modified to introduce pin defects.
For comparison, we used the public data set hosted by Roboflow introduced in Section \ref{sec:related-work}. The dataset contains 198 images showing pin nests only. The dataset is accompanied by annotations for instance segmentation with one class for correct pins and one class for defect pins, as well as a split into training, validation and test images. The provided annotations and dataset split were used without modification to ensure reproducibility and comparability with future studies.

\subsection{Semantic Segmentation}
Six images labelled as pass and six fail images form the training data, with pin contours manually annotated using polygon masks. Since each image contains more than 300 pins, the resulting training set comprises several thousand annotated pin instances. We split images and masks into patches, which results in 576 training images and masks, respectively. We selected U-Net as a well-established semantic segmentation architecture~\cite{unet} with a pre-trained ResNet34 backbone as a compromise between segmentation performance and computational complexity compared to architectures employing deeper ResNet50 backbones.
It was trained for 25 epochs, as preliminary experiments indicated convergence of the validation metrics within this range. 
Additionally, DeepLab~\cite{deeplab}, Fully Convolutional Network (FCN)~\cite{fcn} and Lite Reduced Atrous Spatial Pyramid Pooling (LRASPP)~\cite{lraspp} models are trained as comparison basis. U-Net and DeepLab have been previously evaluated for PCB-mounted component segmentation~\cite{pasunuri}. We selected FCN as a classical semantic segmentation architecture with a comparatively simple decoder structure, while LRASPP was included as a lightweight segmentation model designed for computational efficiency. To prioritise the detection of defective pins, all models were optimised with respect to recall. 

\subsection{Contour Analysis}
\label{sec:contour}
We use thresholding~\cite{opencv} to find contours from the predicted masks of images that were not used in training. The area of each contour was calculated as well as the aspect ratio of the surrounding bounding box. As the area of misaligned pins is typically larger than the area of correctly inserted pins, we use area-related features to train a decision function: average contour area, standard deviation of the contour area, maximum contour area and mean contour aspect ratio. In addition, we included the number of found contours. This feature was originally intended to support a pin-type-dependent analysis. Although pin-type information was not available for both datasets, the number of detected contours was retained as a potentially informative descriptor.

\subsection{Logistic Regression}
\label{sec:log}
We use these features to train a logistic function that decides whether a board is pass or fail based on the board features. Since defective and non-defective boards could not be reliably separated by individual feature thresholds, we employed the classifier to combine the information provided by multiple features. Precision, recall and F1-score were calculated for classes \enquote{pass} and \enquote{fail}. The logistic regression model outputs a probability of a board belonging to the pass class. Model performance was primarily evaluated using threshold-independent metrics, such as ROC-AUC.
For deployment, a decision threshold can be selected depending on the desired trade-off between pass precision and fail recall. Therefore, the decision threshold was varied and evaluated according to the following priorities:
\begin{enumerate}
	\item Maximising pass precision so that fail boards do not pass the quality check.
	\item Maximising fail recall so that no failed board is overlooked.
	\item Maximising fail precision to avoid getting too many fail predictions
	\item Maximising pass recall, which denotes the fraction of boards correctly classified as pass among all actually pass boards.
\end{enumerate}

\subsection{Experimental Overview}
The conducted experiments are summarised as follows. The first two experiments evaluate the proposed method, whereas the latter two provide results for alternative inspection approaches.
\begin{enumerate}
    \item Evaluation of multiple semantic segmentation architectures on the industrial dataset.
    \item Evaluation of the proposed contour-based board-level classification pipeline.
    \item Evaluation of an anomaly-detection baseline.
    \item Evaluation of YOLO instance-segmentation and detection baselines.
\end{enumerate}

\section{Baseline and Comparison Methods}

Since the literature provides limited benchmark results for PCB pin inspection, we additionally evaluated two alternative approaches for automated PCB pin inspection to contextualise the performance of the developed method. We selected PatchCore~\cite{patchcore} as a representative anomaly-detection approach. In contrast to the proposed supervised segmentation workflow, PatchCore can be trained using only defect-free samples and does not require pixel-level annotations. We selected YOLO instance segmentation because the public Roboflow data set provided annotations in the corresponding format and because instance segmentation represents an alternative object-centric approach to pin inspection.

\subsection{PatchCore-Based Anomaly Detection}
\label{sec:patchcore}

PatchCore is a memory-bank-based anomaly detection method for industrial visual 
inspection~\cite{patchcore}. It is trained using non-defective samples only.
A pre-trained CNN extracts local feature
representations from the training images, which are stored in a representative data base. During inference, the local features of a test image are compared
to the stored non-defective features. Large distances to the nearest non-defective features
indicate visually anomalous image regions and result in a high image-level
anomaly score.

For the industrial data, we used YOLOv8-based localisation to extract
pin-nest crops from full PCB images. Then, we applied PatchCore to these crops
to distinguish non-defective from defective pin nests. Consequently, the crop-level
pipeline depends both on the quality of the preceding localisation step and on
the anomaly decision produced by PatchCore.

We trained PatchCore exclusively on
108 manually segmented non-defective pin-nest crops derived from the industrial data set. The evaluation set consisted of
148 non-defective crops generated by the YOLO-based localisation pipeline and
44 defective candidates. The latter were manually
reviewed to separate actual defective pin nests from localisation failures. This
review identified 29 crops containing pin misalignments and we excluded all invalid and unclear crops from the PatchCore evaluation.
For the Roboflow dataset, we used the predefined image splits.
The PatchCore implementation was based on Anomalib version 2.0.0. We used a pre-trained
WideResNet-50-2 backbone to extract feature representations. 

\subsection{YOLO Instance Segmentation}
\label{sec:yolo-overview}

For the industrial data, we used the same manually annotated segmentation masks as for the semantic segmentation experiments. To enable supervised pin-level classification, the pin annotations on defective
boards were reviewed and misaligned pins were assigned to a respective class. The
resulting reviewed two-class annotations formed the common basis for both the
YOLO instance-segmentation and bounding-box detection.

We divided the full-board images into overlapping patches to preserve the
visual detail of individual pins during YOLO training and excluded empty patches from the generated datasets.

For an instance-segmentation variant, we converted the reviewed polygon
annotations into YOLO segmentation labels. 
The main patch-based segmentation model used the pre-trained
YOLOv8s-seg architecture. We trained it on \grid{1024}{1024} pixel
patches for 120 epochs. Additionally, we evaluated a larger YOLOv8m-seg model for
comparison. By preserving the polygon representation, the segmentation variant
was intended to retain detailed spatial information about the shape and extent
of individual pins.
For a bounding-box detection variant, each reviewed pin polygon was converted
into a tight rectangular bounding box defined by the minimum and maximum
coordinates of the polygon extent.
The detection model used the pre-trained YOLOv8s architecture and was
trained on the same \grid{1024}{1024} image patches for 120 epochs.

For the Roboflow dataset, the predefined training, validation, and test splits
were retained. We trained a pre-trained YOLOv8s detector for 50 epochs
with an input size of \grid{1024}{1024} pixels.

The dataset was strongly class-imbalanced, and the test split contained only a single image with defective pins. Consequently, the experiment was conducted primarily to assess the feasibility of a direct YOLO-based inspection pipeline. Due to the limited number of defective test samples, the resulting performance metrics should be interpreted with appropriate caution.


\section{Results}
\label{sec:results}

\begin{table*}[htbp] 
\caption{Segmentation Performance of the Proposed Method on the Industrial and Roboflow Datasets. Best Values on the Industrial Dataset are Highlighted in Bold} 
\label{tab:modeleva} 
\centering 
\begin{tabular}{llccccc} 
\toprule 
Dataset & Model & Accuracy & Precision & Recall & F1-Score & IoU \\
\midrule
\multirow{4}{*}{Industrial}
& U-Net (ResNet34) & \textbf{0.995} & \textbf{0.864} & 0.724 & \textbf{0.788} & \textbf{0.650}\\
& DeepLabV3 (ResNet50) & 0.994 & 0.748 & \textbf{0.817} & 0.781 & 0.640 \\
& FCN (ResNet50) & 0.994 & 0.752 & 0.795 & 0.773 & 0.630 \\
& LRASPP (MobileNetV3) & 0.993 & 0.742 & 0.759 & 0.750 & 0.600 \\
\midrule
Roboflow
& U-Net & 0.981 & 0.936 & 0.767 & 0.843 & 0.728 \\
\bottomrule 
\end{tabular} 
\end{table*}

\begin{table}[htbp]
	\caption{Classification Results of the Proposed Approach and the PatchCore Comparison Method on the Industrial and Roboflow Datasets. Best Values per Metric and Class are Highlighted in Bold}
    \label{tab:log_score}
	\centering
		\begin{tabular}{llcccc}
			\toprule
			 Method & Class & Precision & Recall & F1-Score & Support \\
             \midrule
             \multicolumn{6}{c}{\textbf{Industrial Dataset}} \\
             \midrule

            \multirow{2}{*}{Proposed}
			& pass & \textbf{1.000} & \textbf{0.953} & \textbf{0.976} & 190 \\
            
			& fail & \textbf{0.654} & \textbf{1.000} & \textbf{0.791} & 17 \\
            \cmidrule(lr){2-6}

            \multirow{2}{*}{PatchCore}
            & pass & 0.976 & 0.811 & 0.886 & 148 \\
            
            & fail & 0.481 & 0.897 & 0.626 & 29 \\
            \midrule

            \multicolumn{6}{c}{\textbf{Roboflow Dataset}} \\ 
            \midrule
            \multirow{2}{*}{Proposed}
            & pass & \textbf{1.000} & \textbf{0.970} & \textbf{0.985} & 33 \\ 
        
            & fail & 0.667 & \textbf{1.000} & 0.800 & 2 \\
            \cmidrule(lr){2-6}
            \multirow{2}{*}{PatchCore}
            &pass & 0.935 & 0.967 & 0.951 & 30 \\
            
            &fail & \textbf{0.875} & 0.778 & \textbf{0.824} & 9 \\            
            \bottomrule
		\end{tabular}
\end{table}

\begin{table}[htbp]
    \caption{YOLO-Based Baseline Results on the Industrial and Roboflow Datasets}
    \label{tab:Roboflow-yolo-results}
    \centering
    \begin{tabular}{llcccc}
    \toprule
        Method & Class & Prec. & Rec. & mAP$_{50}$ & mAP$_{50:95}$ \\
        \midrule
        \multicolumn{6}{c}{\textbf{Industrial Dataset}} \\ 
        \midrule
        \multirow{3}{*}{Instance Seg.}
        & All & 0.492 & 0.692 & 0.584 & 0.293 \\
        & pin & 0.787 & 0.848 & 0.784 & 0.319 \\
        & pin\_tilt & 0.197 & 0.535 & 0.384 & 0.268 \\
        \cmidrule(lr){1-6}

        \multirow{3}{*}{BBox Detect.}
        & All & 0.566 & 0.692 & 0.678 & 0.442 \\
        & pin & 0.841 & 0.891 & 0.867 & 0.532 \\
        & pin\_tilt & 0.292 & 0.493 & 0.488 & 0.351 \\
        \midrule

        \multicolumn{6}{c}{\textbf{Roboflow Dataset}} \\
        \midrule
        \multirow{3}{*}{BBox Detect.}
        & All & 0.952 & 0.979 & 0.987 & 0.747 \\
        & pin & 0.999 & 0.989 & 0.995 & 0.861 \\
        & pin\_tilt & 0.905 & 0.970 & 0.979 & 0.634 \\
        \bottomrule
    \end{tabular}
\end{table}

\subsection{Proposed Method Results}
The evaluation of different semantic segmentation models was assessed using accuracy, precision, recall, F1-score, and intersection over union (IoU) and shows best results for the U-Net model, as seen in Table \ref{tab:modeleva}. Only for the recall-value, DeepLabV3 performs superior. Since U-Net achieved the best overall segmentation performance, we selected it for the subsequent feature extraction and board-level classification steps.

\begin{figure}[bp]
	\centerline{\includegraphics[width=0.5\textwidth]{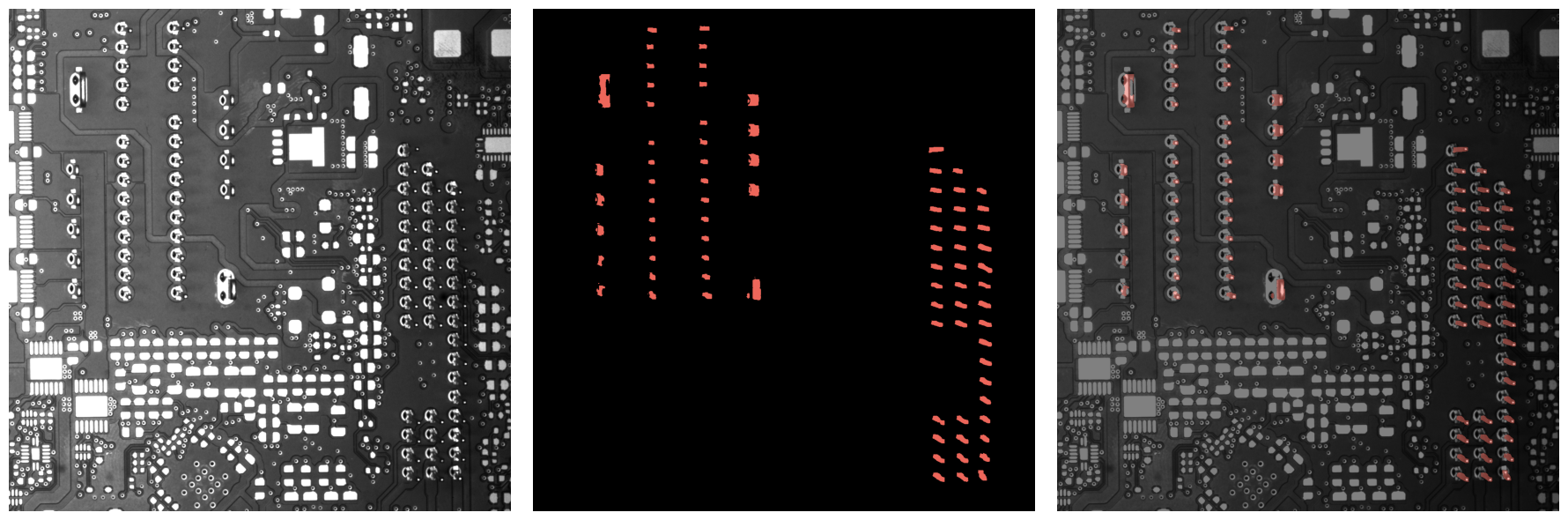}}
	\caption{A section of an industrial test image with the predicted mask and overlay.}
	\label{fig:overlay}
\end{figure}

Fig. \ref{fig:overlay} shows an example of an PCB image detail with its predicted mask by the U-Net model.

The resulting segmentation masks enabled reliable contour extraction using the thresholding procedure described in Section \ref{sec:contour}. The extracted contour features were subsequently used to train the logistic regression classifier.
Training of the logistic decision function achieves an Area Under the Receiver Operating Characteristic Curve (ROC-AUC) of 0.990 and the impact of the features could be measured based on their coefficients:

\begin{enumerate}
	\item average area (-1.862)
	\item area standard deviation (-1.406)
	\item maximum area (-0.934)
	\item mean aspect ratio (-0.314)
	\item number of contours (-0.100)
\end{enumerate}

To illustrate the separation between pass and fail boards, a decision threshold of 0.430 was selected based on the threshold analysis described in Section~\ref{sec:log}.

\begin{figure}[htbp]
	\centerline{\includegraphics{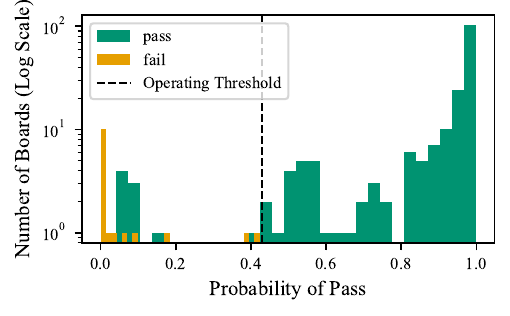}}
	\caption{Distribution of the predicted pass probability for pass and fail boards. The dashed line indicates the operating threshold used for board-level classification. A logarithmic y-axis is used to improve the visibility of fail boards. The figure illustrates the observed separation between pass and fail boards achieved by the proposed method.}
	\label{fig:logreg}
\end{figure}

Fig. \ref{fig:logreg} shows that fail boards are assigned lower values of pass probabilities, whereas pass boards are concentrated at higher probabilities. 
Using the selected operating threshold, all fail boards are separated from the majority of pass boards. The resulting classification metrics are reported in Table~\ref{tab:log_score}.

\begin{figure}[tp]
	\centerline{\includegraphics[width=0.45\textwidth]{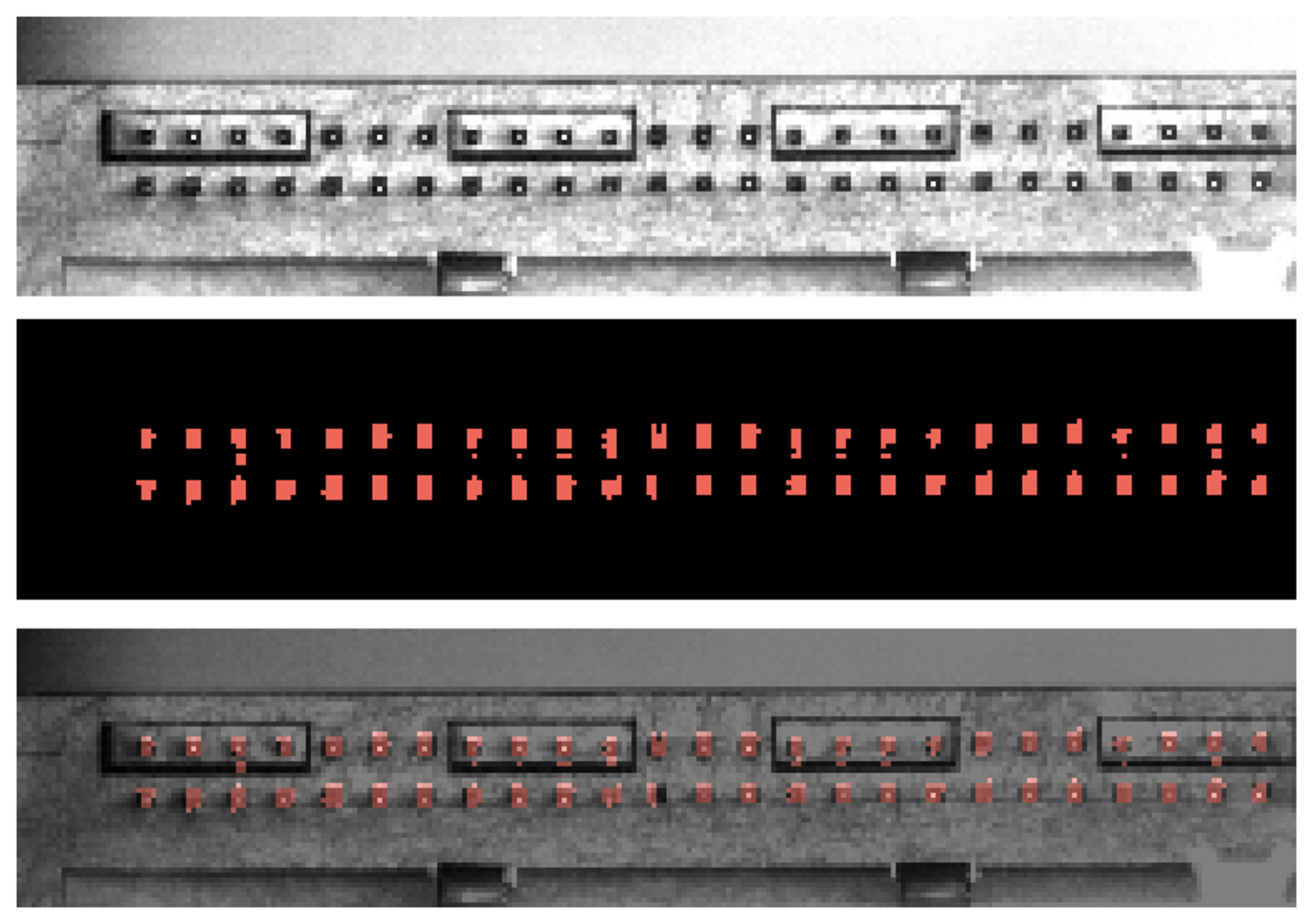}}
	\caption{Original image, predicted mask and overlay for an example of the Roboflow data.}
	\label{fig:robo}
\end{figure}

We tested the whole pipeline for the Roboflow dataset, achieving similar results for the segmentation as seen in Table~\ref{tab:modeleva} and depicted in Fig. \ref{fig:robo}.

Logistic regression training resulted in new coefficients, with most impact given by the maximum contour (-2.530) followed by the number of contours (1.253) and least impact by mean aspect ratio (-0.047). The logistic model achieved a 1.000 ROC-AUC with metrics depicted in Table \ref{tab:log_score}.

Runtime measurements were performed on a server equipped with an NVIDIA A100-SXM4-40GB GPU,  AMD EPYC 7662 64-Core processor, and \SI{1}{\tera\byte} RAM. Processing a \grid{4096}{3000} board image required approximately \SI{18.0}{\second} on average, including semantic segmentation, contour extraction, feature computation, and board-level classification. Semantic segmentation accounted for the majority of the processing time, while feature extraction and classification required approximately \SI{0.23}{\second} per board.

\subsection{PatchCore Baseline Results}
\label{sec:patchcore-results}

Table~\ref{tab:log_score} summarises the evaluated PatchCore metrics on the crop test set and the Roboflow test set, which is on pin nest-level for both datasets. The selected configuration used the WRN50-2 backbone.

\subsection{YOLO Baseline Results}
\label{sec:Roboflow-results}

Table~\ref{tab:Roboflow-yolo-results} summarises the YOLO detection performance
on the Roboflow validation and test splits. Class-wise results are
reported in addition to the overall metrics in order to make the effect of the
strong class imbalance visible. 
Table~\ref{tab:Roboflow-yolo-results} reports the test performance of the patch-based YOLOv8s-seg model using the default confidence threshold.

As shown in Table~\ref{tab:Roboflow-yolo-results}, correct pins were
segmented substantially more reliably than tilted pins.

\section{Discussion}

Although previous research stated only moderate suitability of U-Net for PCB component segmentation compared to DeepLab \cite{pasunuri}, it achieved a high IoU score for pin segmentation for both datasets. Nevertheless, DeepLab achieved a better recall value, indicating better suitability for detecting pins than the overall more precise U-Net. Semantic pin segmentation achieved high precision scores, and the method can handle large image sizes because of the possibility of cropping them into patches without problems regarding annotations. 
The selection of features to train a logistic model resulted in optimal values for pass precision and fail recall, which we gave highest priority. It is notable that the coefficients changed substantially for logistic regression trained on the Roboflow data. For this data, the maximum contour area and number of contours had larger absolute values. This could be caused by overlapping contours, as tilted pins get larger bounding boxes. As the scores for the Roboflow data are slightly higher than for the industrial data, bounding box annotations may be more suitable than polygon masks. 
Overall, we developed the method for an industrial dataset, but achieved similarly promising results on publicly available data. Both datasets are homogeneous, but the difference between the two data sets is quite large regarding both images and masks, as one set contains fully depicted PCBs with polygon masks and the other contains cropped pin nests with annotated bounding boxes. This indicates that the method is robust to domain changes given that the learning functions are retrained.
Because of the homogeneity of the two used datasets, no statement can be made for highly heterogeneous datasets as semantic segmentation and logistic regression training may have overfitted. A detailed analysis of detection performance as a function of the number of defective pins per board was not possible for the available datasets. In the industrial dataset, defect boards typically contain multiple manually introduced pin defects and the exact number of affected pins is not annotated. In contrast, the public dataset mainly contains images with few defective pins but provides only a limited number of defect samples. Consequently, the available data do not support a statistically meaningful analysis across different defect counts. This constitutes an interesting direction for future work.

The investigated alternative inspection approaches provide additional evidence that the developed feature-based method is well-performing regarding PCB quality assurance based on pin tilting defects. The performance of the baseline YOLO instance segmentation is more closely tied to the used data and annotations, as the results were excellent on the Roboflow data but not promising on the industrial data.
The PatchCore anomaly detection baseline achieved promising results on both datasets and needs less effort in data labelling compared to the segmentation methods, which is time-consuming and error-prone due to light reflections and image resolutions. While PatchCore supports anomaly localisation through anomaly heatmaps, it does not provide additional information about board-specific features, which is a clear advantage of the proposed method combining semantic segmentation with a feature analysis.
The training and evaluation of the three methods used different data because of different labelling requirements and model purpose: our proposed method works on board-level, the PatchCore baseline on nest-level and the YOLO instance segmentation baseline on pin-level. Although the public Roboflow data set remained the same for all methods, the results are not fully conclusive due to class imbalance.
In this work, we focused on developing a board-level decision function. The investigated alternative approaches were therefore intended as complementary points of reference rather than as directly comparable benchmark methods. A more comprehensive comparative study could provide further insights in the future.

\section{Conclusion}

We developed a method for pin insertion defect detection focusing on misaligned pins, which is based on semantic segmentation using a U-Net model, with subsequent contour detection and feature extraction to train a logistic classifier. We trained and tested the method on industrial and publicly available image data. The resulting decision function classifies pass and fail on board-level and achieved 0.990 ROC-AUC and 1.000 ROC-AUC for the test datasets, indicating a strong separation. While these results demonstrate the potential of the approach, the observed performance should be interpreted in the context of the available datasets and their respective class distributions. In particular, the used public test set contains only a small number of defective boards, limiting the statistical significance of the reported fail-class performance.
For further investigation, we tested the anomaly detection method PatchCore on pin nest-level and YOLO instance segmentation on pin-level. While instance segmentation showed large differences between both datasets, PatchCore achieved precision and recall values comparable to those of the proposed method. We conclude that PatchCore benefits from not requiring manual annotations. However, it does not explicitly incorporate PCB-specific information. In contrast, the proposed feature-based method exploits PCB-specific geometric information but requires pixel-level annotations for semantic segmentation. The compared approaches operate at different levels of granularity, limiting direct comparability. We therefore recommend to investigate the difference in board-, nest- and pin-level methods more comprehensively in future work.

\section*{Acknowledgements}

We acknowledge partial support by the European Regional Development Fund (ERDF) and by the Free State of Bavaria as part of the project AIM-SMEs (Grant No. 2506-014-3.2), co-funded by the European Union; also by the High-Tech Agenda of the Free State of Bavaria.
We thank GEFASOFT Automatisierung und Software GmbH for providing the image data used in this work, and Denis Bucher for technical support.

\bibliographystyle{IEEEtran}
\bibliography{references}

\end{document}